\def\year{2016}\relax
\documentclass[letterpaper]{article}
\usepackage{aaai16}
\usepackage{times}
\usepackage{helvet}
\usepackage{courier}
\usepackage{graphicx}
\usepackage{amsmath}
\usepackage{amssymb}
\usepackage{amsthm}
\usepackage{algorithm}
\usepackage{algorithmic}
\usepackage{tabu}
\usepackage{subfigure}
\usepackage{wrapfig}

\newcommand{\prob}{\mathcal{P}}
\newcommand{\ep}{\mathbb{E}}
\newcommand{\lambdav}{\boldsymbol \lambda }
\def\indicator{{\mathbb I}}
\newtheorem{theorem}{{\bf Theorem}}
\newtheorem{lemma}[theorem]{{\bf Lemma}}

\begin{document}
%
\title{Discriminative Nonparametric Latent Feature Relational Models \\with Data Augmentation}
\author{Bei Chen$^\dagger$, Ning Chen$^{\ddagger}$\thanks{Corresponding authors.}, Jun Zhu$^{\dagger *}$, Jiaming Song$^\dagger$, Bo Zhang$^\dagger$\\
$^\dagger$Dept. of Comp. Sci. \& Tech., State Key Lab of Intell. Tech. \& Sys., Center for Bio-Inspired Computing Research,\\
$^\ddagger$MOE Key lab of Bioinformatics, Bioinformatics Division and Center for Synthetic \& Systems Biology,\\
TNList, Tsinghua University, Beijing, 100084, China\\
\{chenbei12@mails., ningchen@, dcszj@, sjm12@mails., dcszb@\}tsinghua.edu.cn\\
}
\maketitle
\begin{abstract}
We present a discriminative nonparametric latent feature relational model (LFRM) for link prediction to automatically infer the dimensionality of latent features. Under the generic RegBayes (regularized Bayesian inference) framework, we handily incorporate the prediction loss with probabilistic inference of a Bayesian model; set distinct regularization parameters for different types of links to handle the imbalance issue in real networks; and unify the analysis of both the smooth logistic log-loss and the piecewise linear hinge loss. For the nonconjugate posterior inference, we present a simple Gibbs sampler via data augmentation, without making restricting assumptions as done in variational methods. We further develop an approximate sampler using stochastic gradient Langevin dynamics to handle large networks with hundreds of thousands of entities and millions of links, orders of magnitude larger than what existing LFRM models can process. Extensive studies on various real networks show promising performance.
\end{abstract}

\section{Introduction}
Link prediction is a fundamental task in statistical network analysis. For static networks, it is defined as predicting the missing links from a partially observed network topology (and some attributes if exist). Existing approaches include: 1) Unsupervised methods that design good proximity/similarity measures between nodes based on network topology features~\cite{Liben-Nowell03}, e.g., common neighbors, Jaccard's coefficient~\cite{Salton:1983}, Adamic/Adar~\cite{Adamic:2003}, etc; 2) Supervised methods that learn classifiers on labeled data with a set of manually designed features~\cite{lichtenwalter2010,al2006,Shi:2009}; 3) others \cite{backstrom2011} that use random walks to combine the network structure information with node and edge attributes. One possible limitation for such methods is that they rely on well-designed features or measures, which can be time demanding to get and/or application specific.

Latent variable models \cite{hoff2002,hoff2007,Chang:RTM09} have been widely applied to discover latent structures from complex network data, based on which prediction models are developed for link prediction. Although these models work well, one remaining problem is how to determine the unknown number of latent classes or features. A typical way using model selection, e.g., cross-validation or likelihood ratio test \cite{LiuShao:03}, can be computationally prohibitive by comparing many candidate models. Bayesian nonparametrics has shown promise in bypassing model selection by imposing an appropriate stochastic process prior on a rich class of models~\cite{Antoniak:74,Griffiths05}. For link prediction, the infinite relational model (IRM) \cite{kemp2006} is class-based and uses Bayesian nonparametrics to discover systems of related concepts. One extension is the mixed membership stochastic blockmodel (MMSB) \cite{Airoldi08}, which allows entities to have mixed membership. \cite{Miller09} and \cite{Zhu12} developed nonparametric latent feature relational models (LFRM) by incorporating Indian Buffet Process (IBP) prior to resolve the unknown dimension of a latent feature space. Though LFRM has achieved promising results, exact inference is intractable due to the non-conjugacy of the prior and link likelihood. One has to use Metropolis-Hastings \cite{Miller09}, which may have low accept rates if the proposal distribution is not well designed, or variational inference \cite{Zhu12} with truncated mean-field assumptions, which may be too strict in practice.

In this paper, we develop discriminative nonparametric latent feature relational models (DLFRM) by exploiting the ideas of data augmentation with simpler Gibbs sampling \cite{Polson2011,Polson2013} under the regularized Bayesian inference (RegBayes) framework \cite{Zhu:JMLR2014_2}. Our major contributions are:
1) We use the RegBayes framework for DLFRM to deal with the imbalance issue in real networks and naturally analyze both the logistic log-loss and the max-margin hinge loss under a unified setting; 2) We explore data augmentation techniques to develop a simple Gibbs sampling algorithm, which is free from unnecessary truncation and assumptions that typically exist in variational approximation methods; 3) We develop an approximate Gibbs sampler using stochastic gradient Langevin dynamics, which can handle large networks with hundreds of thousands of entities and millions of links (See Table~\ref{table:data}), orders of magnitude larger than what the existing LFRM models~\cite{Miller09,Zhu12} can process; and 4) Finally, we conduct experimental studies on a wide range of real networks and the results demonstrate promising results of our methods.

\section{Nonparametric LFRM Models}

We consider static networks with $N$ entities. Let $Y$ be the $N \times N$ binary link indicator matrix, where $y_{ij}=1$ denotes the existence of a link from entity $i$ to $j$, and $y_{ij}=-1$ denotes no link from $i$ to $j$. $Y$ is not fully observed.
\begin{wrapfigure}{r}{0.18\textwidth}
\includegraphics[width=0.18\textwidth]{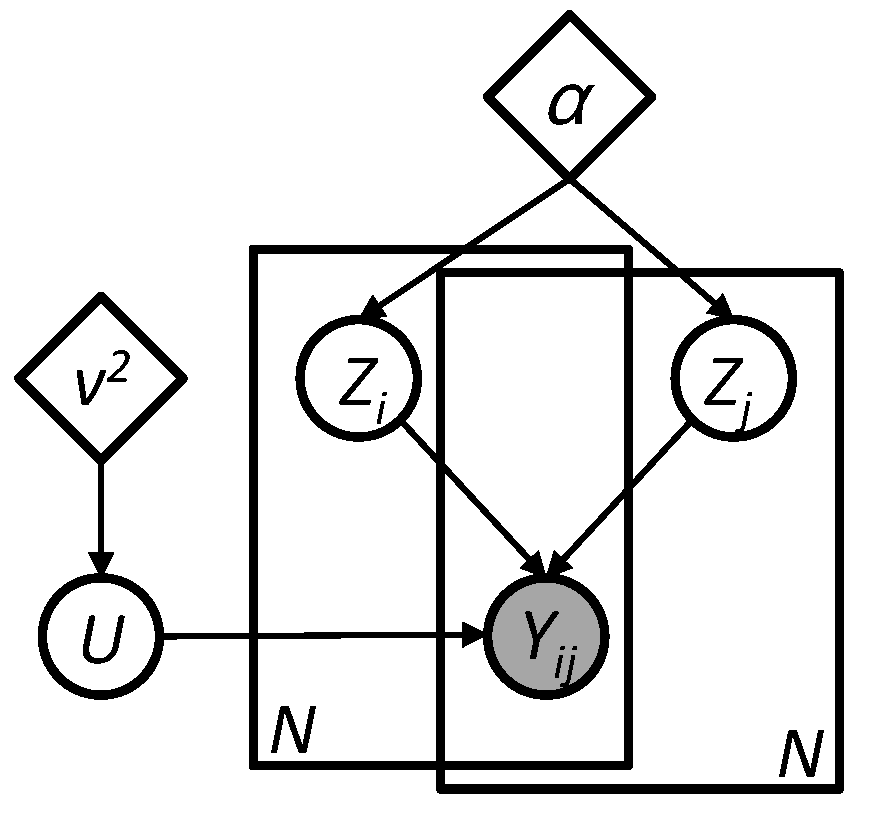}
\caption{The graphical structure of LFRM.}\label{fig:model}
\end{wrapfigure}
Our goal is to learn a model from the partially observed links and predict the values of the unobserved entries of $Y$. Fig.~\ref{fig:model} illustrates a latent feature relational model (LFRM), where each entity is represented by $K$ latent features. Let $Z$ be the $N\times K$ feature matrix, each row is associated with an entity and each column corresponds to a feature.
We consider the binary features\footnote{Real-valued features can be learned by using composition, e.g., $R_i = Z_i \otimes H_i$, where $H_i$ is the real-valued vector representing the amplitudes of each feature while the binary vector $Z_i$ represents the presence of each feature.}: If entity $i$ has feature $k$, then $z_{ik}=1$, otherwise $z_{ik}=0$. Let $Z_i$ be the feature vector of entity $i$, $U$ be a $K \times K$ real-valued weight matrix, $\eta = \textrm{vec}(U)$ and $Z_{ij}=\textrm{vec}(Z_i^\top Z_j)$, where $\textrm{vec}(A)$ is a vector concatenating the row vectors of matrix $A$. Note that $\eta$ and $Z_{ij}$ are column vectors, while $Z_i$ is a row vector. Then the probability of the link from entity $i$ to $j$ is
\begin{equation}\label{eq:likelihood}
p(y_{ij}=1|Z_i, Z_j, U)=\sigma\left(Z_i U Z_j^\top\right)= \sigma\left(\eta^\top Z_{ij} \right),
\end{equation}where{\small{ $\sigma(x)\!\!    =\! \!   \frac{1}{1+\exp(-x)}$}} is the sigmoid function. We assume that links are conditionally independent given $Z$ and $U$, then the link likelihood is{\small{  $p(Y|Z, U) = \prod_{(i, j)\in \mathcal{I}} p(y_{ij}|Z_i, Z_j, U)$}}, where $\mathcal{I}$ is the set of training links (observed links).

In the above formulation, we assume that the dimensionality of the latent features $K$ is known a priori. However, this assumption is often unrealistic especially when dealing with large-scale applications. The conventional approaches that usually need a model selection procedure (e.g., cross validation) to choose an appropriate value by trying on a large set of candidates can be expensive and often require extensive human efforts on guiding the search. Recent progress on Bayesian optimization~\cite{Snoek:2012} provides more effective solution to searching for good parameters, but still needs to learn many models under different configurations of the hyper-parameter $K$.

In this paper, we focus on the nonparametric Bayesian methods~\cite{Griffiths05} for link prediction. The recently developed nonparametric latent feature relational models (LFRM) \cite{Miller09} leverage the advancement of Bayesian nonparametric methods to automatically resolve the unknown dimensionality of the feature space by applying a flexible nonparametric prior. It assumes that each entity $i$ has an infinite number of binary features, that is $Z_i \in \{0, 1\}^\infty$, and the Indian Buffet Process (IBP) \cite{Griffiths05} is used as a prior of $Z$ to produce a sparse latent feature vector for each entity.

We treat the weight matrix $U$ as random and put a prior on it for fully Bayesian inference. Then with Bayes' theorem, the posterior distribution is
\begin{equation}\label{e2}
q(Z, U | Y) \propto p_0(Z)p_0(U)p(Y|Z, U),
\end{equation}where the prior $p_0(Z)$ is an IBP and $p_0(U)$ is often assumed to be an isotropic Gaussian prior.

\subsection{Discriminative LFRM Models}

The conventional Bayesian inference as above relies on Bayes' rule to infer the posterior distribution. In fact, this procedure can be equivalently formulated as solving an optimization problem. For example, the Bayes posterior in Eq. (\ref{e2}) is equivalent to the solution of the following problem:
\begin{eqnarray}
\min_{q(Z, U)\in \prob} { \!\!\rm{KL}} (q(Z, U)||p_0(Z, U))\! - \!\ep_q[\log p(Y|Z, U)],
\end{eqnarray}where $\prob$ is the space of well-defined distributions and ${ \rm{KL}} (q||p)$ is the Kullback-Leibler (KL) divergence from $q$ to $p$. Such an optimization view has inspired the development of regularized Bayesian inference (RegBayes) which solves:
\begin{eqnarray}\label{ee3}
\min_{q(Z, U)\in \prob} { \rm{KL}} (q(Z, U)||p_0(Z, U)) + c \cdot \mathcal{R}( q(Z, U) ),
\end{eqnarray}where $\mathcal{R}(q)$ is a posterior regularization defined on the target posterior distribution and $c$ is a non-negative regularization parameter that balances the prior part and the posterior regularization part. We refer the readers to \cite{Zhu:JMLR2014_2} for more details on a generic representation theorem of the solution and its application~\cite{Zhu:JMLR2014,Mei:robustBayes14} to learn latent feature models for classification. Below, we explore the ideas to develop effective latent feature relational models for link prediction.

Although we could define an averaging classifier and make predictions using the sign rule $\hat{y}_{ij} \!\!= \!\!\textrm{sign} ( \ep_q[ Z_i U Z_j^\top \!] )$, the resulting problem needs to be approximately solved by truncated variational methods, which can be inaccurate in practice. Here, we propose to define a Gibbs classifier, which admits simple and efficient sampling algorithms that are guaranteed to be accurate. Our Gibbs sampler randomly draws the latent variables $(Z,U)$ from the unknown but pre-assumed to be given posterior distribution $q(Z, U)$. Once $Z$ and $U$ are given, we can make predictions using the sign rule $\hat{y}_{ij} \!\!= \!\!\textrm{sign}( Z_i U Z_j^\top  )$ and measure the training error $r(Z,U) = \sum_{(i,j) \in \mathcal{I}} \indicator(y_{ij} \neq \hat{y}_{ij})$, where $\indicator(\cdot)$ is an indicator function. Since the training error is non-smooth and non-convex, it is often relaxed by a well-behaved loss function. Let $\omega_{ij} = Z_i U Z_j^\top$, two well-studied examples are the logistic log-loss $r_1$ and the hinge loss $r_2$:
{\small{\begin{eqnarray}
r_1(Z, U) &=& - \sum_{(i,j) \in \mathcal{I}} \log p(\tilde{y}_{ij} | Z_i, Z_j, U), \nonumber\\
r_2(Z, U) &=& \sum_{(i,j) \in \mathcal{I}} (\ell - {y}_{ij} \omega_{ij})_+, \nonumber
\end{eqnarray}}}where $p(\tilde{y}_{ij}|Z_i, Z_j, U)=\frac{e^{\omega_{ij}\tilde{y}_{ij}}}{1+e^{\omega_{ij}}}$,  $(x)_+ := \max(0,x)$, $\ell$ is the pre-defined cost to penalize a wrong prediction, and $\tilde{y}_{ij}=(y_{ij}+1)/2$ so that $0$ refers to a negative link instead of $-1$. To account for the uncertainty of the latent variables, we define the posterior regularization as the expected loss:
\begin{eqnarray}
\mathcal{R}_1\! (q(U, Z))\! = \!\ep_q[ r_1(Z, U) ] \nonumber, \mathcal{R}_2\! (q(U, Z)) \!= \! \ep_q[ r_2(Z, U) ]. \!\!\!\!\!\!\!\!\!\!\!\!\!\nonumber
\end{eqnarray}With these posterior regularization functions, we can do the RegBayes as in problem~(\ref{ee3}), where the parameter $c$ balances the influence between the prior distribution (i.e., $\mathrm{KL}$ divergence) and the observed link structure (i.e., the loss term). We define the un-normalized pseudo link likelihood:
\begin{eqnarray}
\varphi_1(\tilde{y}_{ij}|Z_i, Z_j, U) &=& \frac{(e^{\omega_{ij}})^{c\tilde{y}_{ij}}}{(1+e^{\omega_{ij}})^{c}}, \label{fai1}\\
\varphi_2({y}_{ij}|Z_i, Z_j, U) &=& \exp(-2c (\ell-{y}_{ij}\omega_{ij})_+ ). \label{fai2}
\end{eqnarray}Then problem~(\ref{ee3}) can be written in the equivalent form:
\begin{equation}\label{ee6}
\min_{q(Z, U)\in \mathcal{P}} { \!\!\!\rm{KL}} (q(Z, U)||p_0(Z, U)) - \mathbb{E}_q[\log \varphi(Y|Z, U)],
\end{equation}where $\varphi(Y|Z, U)=\prod_{i,j \in \mathcal{I}} \varphi(y_{ij}|Z_i, Z_j, U)$ and $\varphi$ can be $\varphi_1$ or $\varphi_2$. Then the optimal solution of (\ref{ee3}) or (\ref{ee6}) is the following posterior distribution with link likelihood:
\begin{eqnarray}\label{ee7}
q(Z, U|Y) \propto p_0(Z)p_0(U)\varphi(Y|Z, U).
\end{eqnarray}Notice that if adopting the logistic log-loss, we actually obtain a generalized pseudo-likelihood which is a powered form of likelihood in Eq. (\ref{eq:likelihood}).

For real networks, positive links are often highly sparse as shown in Table~\ref{table:data}. Such sparsity could lead to serious imbalance issues in supervised learning, where the negative examples are much more than positive examples. In order to deal with the imbalance issue in network data and make the model more flexible, we perform RegBayes by controlling the regularization parameter. For example, we can choose a larger $c$ value for the fewer positive links and a relatively smaller $c$ for the larger negative links. This strategy has shown effective in dealing with imbalanced data in~\cite{Chen13,Zhu12}. We will provide experiments to demonstrate the benefits of RegBayes on dealing with imbalanced networks when learning nonparametric LFRMs.

\section{Gibbs Sampling with Data Augmentation}

As we do not have a conjugate prior on $U$, exact posterior inference is intractable. Previous inference methods for nonparametric LFRM use either Metropolis-Hastings~\cite{Miller09} or variational techniques~\cite{Zhu12} which can be either inefficient or too strict in practice. We explore the ideas of data augmentation to give the pseudo-likelihood a proper design, so that we can directly obtain posterior distributions and develop efficient Gibbs sampling algorithms. Specifically, our algorithm relies on the following unified representation lemma.

\begin{lemma}
Both $\varphi_1$ and $\varphi_2$ can be represented as
{\small{\begin{eqnarray}
\varphi(y_{ij}|Z_i,Z_j,U) \propto \int_0^{\infty}\!\!\exp{\Big(\!\kappa_{ij}\omega_{ij}\!-\!\frac{ \rho_{ij} \omega_{ij}^2}{2} \Big)}\!\phi(\lambda_{ij}) {\rm{d}} \lambda_{ij},\nonumber
\end{eqnarray}}}
where for $\varphi_1$ we have
{\small{$$\kappa_{ij}=c(\tilde{y}_{ij}-\frac{1}{2}),~\rho_{ij} = \lambda_{ij}, ~ \phi(\lambda_{ij})=\mathcal{PG}(\lambda_{ij};c, 0);$$}} while for $\varphi_2$, let $\gamma_{ij} = \lambda_{ij}^{-1}$, we have
{\small{$$\kappa_{ij}=c{y}_{ij}(1 + c \ell  \gamma_{ij}),~ \rho_{ij} = c^2 \gamma_{ij},~ \phi(\lambda_{ij})=\mathcal{GIG}(\frac{1}{2}, 1, c^2\ell^2).$$}}
\end{lemma}

We have used $\mathcal{PG}$ to denote a Polya-Gamma distribution~\cite{Polson2013} and $\mathcal{GIG}$ to denote a generalized inverse Gaussian distribution. We defer the proof to Appendix A\footnote{Supplemental Material:\\http://bigml.cs.tsinghua.edu.cn/\%7Ebeichen/pub/DLFRM2.pdf}, which basically follows~\cite{Polson2013,Polson2011} with some algebraic manipulation on re-organizing the terms.

\begin{algorithm}[tb]
\caption{Gibbs sampler for DLFRM}
\label{alg1}
\begin{algorithmic}
\STATE {\bfseries Init:} draw $Z$ from IBP, $U$ from $\mathcal{N}(0, \nu^{-2})$; set $\lambda=1$.
\FOR {$iter=1, 2, \dots, L$}
\FOR {$n=1, 2, \dots, N$}
\STATE draw $\{z_{nk}\}_{k=1}^K$ from Eq. (\ref{ee8});
\STATE draw $k_n$ using Eq. (\ref{e6}).
\IF{$k_n>0$}
\STATE  update $K \gets K + k_n$, update new weights;
\ENDIF
\ENDFOR
\STATE draw $U$ using Eq. (\ref{e7}) and draw $\lambda$ using Eq. (\ref{e8}).
\ENDFOR
\end{algorithmic}
\end{algorithm}

\subsection{Sampling Algorithm}

Lemma 1 suggests that the pseudo-likelihood $\varphi$ can be considered as the marginal of a higher dimensional distribution that includes the augmented variables $\lambda$:
\begin{eqnarray}
\psi(\lambda, Y|Z, U) \!\propto\!\! \prod_{(i, j)\in \mathcal{I}} \exp{\left( \kappa_{ij}\omega_{ij}-\frac{\rho_{ij}\omega_{ij}^2}{2} \right) } \phi(\lambda_{ij}),
\end{eqnarray}which is a mixture of Gaussian components of $U$ once $Z$ is given, suggesting that we can effectively perform Gibbs sampling if a conjugate Gaussian prior is imposed on $U$. We also construct the complete posterior distribution:
\begin{eqnarray}
q(Z,U,\lambda|Y) \propto p_0(Z)p_0(U)\psi(\lambda,Y|Z, U),
\end{eqnarray}such that our target posterior $q(Z,U|Y)$ is a marginal distribution of the complete posterior. Therefore, if we can draw a set of samples $\{(Z_t, U_t, \lambda_t) \}_{t=1}^L$ from the complete posterior, by dropping the augmented variables, the rest samples $\{(Z_t, U_t) \}_{t=1}^L$ are drawn from the target posterior $q(Z,U | Y)$. This technique allows us to sample the complete posterior via a Gibbs sampling algorithm, as outlined in Alg.~\ref{alg1} and detailed below.

{\bf For $Z$:}
We assume the Indian Buffet Process (IBP) prior on the latent feature $Z$. Although the total number of latent features is infinite, every time we only need to store $K$ active features that are not all zero in the columns of $Z$. When sampling the $n$-th row, we need to consider two cases, due to the nonparametric nature of IBP.

First, for the active features, we sample $z_{nk}(k=1,...,K)$ in succession from the following conditional distribution
\begin{equation}
q(z_{nk}|Z_{-nk},\eta, \lambda) \propto p(z_{nk})\psi(\lambda, Y|Z_{-nk},\eta,z_{nk}), \label{ee8}
\end{equation}where $p(z_{nk}=1) \propto m_{-n,k}$ and $m_{-n,k}$ is the number of entities containing feature $k$ except entity ${n}$.

Second, for the infinite number of remaining all-zero features, we sample $k_n$ number of new features and add them to the $n$th row. Then we get the new $N \times (K+k_n)$ matrix $Z^*$ which becomes old when sampling the $(n+1)$-th row. Every time when the number of features changes, we also update $U$ and extend it to a $(K+k_n)\times(K+k_n)$ matrix $U^*$. Let $Z'$ and $U'$ be the parts of $Z^*$ and $U^*$ that correspond to the $k_n$ new features. Also, we define $\eta'=\textup{vec}(U')$. During implementation, we can delete the all-zero columns after every resampling of $Z$, but here we ignore it. Let $\eta‘$ follow the isotropic Normal prior $\mathcal{N}(0, \nu^{-2})$. Now the conditional distribution for $k_n=0$ is $p(k_n=0|Z, \eta, \lambda) = p_0(k_n)$, and the probability of $k_n\not=0$ is
\begin{eqnarray}\label{e6}
p(k_n\not=0|Z, \eta, \lambda) \! =\! p_0(k_n)|\Sigma|^{\frac{1}{2}}\nu^D \!\exp{\Big( \frac{1}{2}\mu^\top \Sigma^{-1} \mu \Big)},
\end{eqnarray}where $p_0(k_n) = \textup{Poisson}\left( k_n;\frac{\alpha}{N} \right)$ is from the IBP prior, $D=2k_n K+k_n^2$ is the dimension of $\eta'$ and the mean $\mu=\Sigma(\sum_{(i, j)\in \mathcal{I}}(\kappa_{ij} - \rho_{ij}\omega_{ij})Z_{ij}')$ , covariance $\Sigma = (\sum_{(i, j)\in \mathcal{I}}\rho_{ij}Z_{ij}' {Z_{ij}'}^\top+ {\nu^2} I)^{-1}$.

We compute the probabilities for $k_n\!=\!0,1,...,K_{max}$, do normalization and sample from the resulting multinomial. Here, $K_{max}$ is the maximum number of features to add. Once we have added $k_n\!(\not=\!\!\!0)$ new features, we should also sample their weights $\eta'$, which follow a $D$ dimensional multivariate Gaussian, in order to resample the next row of $Z$.

{\bf For $U$:} After the update of $Z$, we resample $U$ given the new $Z$. Let $\tilde{D}=K\times K$ and $\eta$ follow the isotropic Normal prior $p_0(\eta)=\prod_{d=1}^{\tilde{D}}\mathcal{N}(\eta_d; 0, \nu^{-2})$. Then the posterior is also a Gaussian distribution
\begin{eqnarray}\label{e7}
q(\eta |\lambda, Z) \propto p_0(\eta)\psi(\lambda, Y|Z, \eta) = \mathcal {N}(\eta;\tilde{\mu},\tilde{\Sigma}),
\end{eqnarray}with the mean $\tilde{\mu}=\tilde{\Sigma}(\sum_{(i, j)\in \mathcal{I}}\kappa_{ij}Z_{ij})$ and the convariance $\tilde{\Sigma} = (\sum_{(i, j)\in \mathcal{I}}\rho_{ij}Z_{ij}Z_{ij}^\top +  {\nu^2} I )^{-1}$.

{\bf For $\lambdav$:}
Since the auxiliary variables are independent given the new $Z$ and $U$, we can draw each $\lambda_{ij}$ separately. From the unified representation, we have
\begin{eqnarray}\label{e8}
q(\lambda_{ij}|Z,\eta) \propto \exp{\Big( \kappa_{ij} \omega_{ij} - \frac{\rho_{ij} \omega_{ij}^2}{2} \Big)}\phi(\lambda_{ij}).
\end{eqnarray}By doing some algebra, we can get the following equations. For $\varphi_1$, $\lambda_{ij}$ still follows a Polya-Gamma distribution $q(\lambda_{ij}|Z,\eta) = \mathcal{PG}(\lambda_{ij};c,\omega_{ij})$, from which a sample can be efficiently drawn. For $\varphi_2$, $\lambda_{ij}$ follows a generalized inverse Gaussian distribution $q(\lambda_{ij}|Z,U,Y) = \mathcal{GIG} ( \frac{1}{2},1,c^2\zeta_{ij}^2 )$, where $\zeta_{ij} = \ell - y_{ij}\omega_{ij}$. Then $\gamma_{ij} := \lambda_{ij}^{-1}$ follows an inverse Gaussian distribution $q(\gamma_{ij} | Z,U,Y)=\mathcal{IG} ( \frac{1}{c|\zeta_{ij}|},1 )$, from which a sample can be easily drawn in a constant time.

\subsection{Stochastic Gradient Langevin Dynamics}
Alg. \ref{alg1} needs to sample from a $K^2$-dim Gaussian distribution to get $U$, where $K$ is the latent feature dimension. This procedure is prohibitively expensive for large networks when $K$ is large (e.g., $K>40$). To address this problem, we employ stochastic gradient Langevin dynamics (SGLD) \cite{welling2011bayesian}, an efficient gradient-based MCMC method that uses unbiased estimates of gradients with random mini-batches. Let $\theta$ denote the model parameters and $p(\theta)$ is a prior distribution. Given a set of i.i.d data points $\mathcal{D}=\{x_i\}_{i=1}^M$, the likelihood is $p(\mathcal{D}|\theta)=\prod_{i=1}^M p(x_i|\theta)$. At each iteration $t$, the update equation for $\theta$ is:
\begin{equation}
\Delta \theta_t\! =\! \frac{\epsilon_t}{2} \Big( \nabla \log p(\theta_t) + \frac{M}{m} \!\!\sum_{x_i\in {\mathcal{D}_t}} \!\! \nabla \log p(x_{i} | \theta_t) \Big) + \delta_t ,
\end{equation}where $\epsilon_t$ is the step size, ${\mathcal{D}_t}$ is a subset of ${\mathcal{D}}$ with size $m$ and $\delta_t \sim \mathcal{N}(0, \epsilon_t)$ is the Gaussian noise. When the stepsize is annealed properly, the Markov chain will converge to the true posterior distribution.

Let ${\mathcal{I}_t}$ be a subset of ${\mathcal{I}}$ with size $m$. We can apply SGLD to sample $\eta$ (i.e., $U$). Specifically, according to the true posterior of $\eta$ as in Eq. (\ref{e7}), the update rule is:
\begin{equation}
\Delta \eta_t\! =\! \frac{\epsilon_t}{2} \Big( -{\nu^2}\eta_t + \frac{|\mathcal{I}|}{m} \!\!\sum_{(i, j)\in {\mathcal{I}_t}} \!\! (\kappa_{ij} - \rho_{ij}\omega_{ij})Z_{ij} \Big) + \delta_t ,
\end{equation}where $\delta_t$ is a $K^2$-dimensional vector and each entry is a Gaussian noise. {After a few iterations, we will get the approximate sampler of $\eta$ (i.e., $U$) very efficiently.}

\begin{table}[t]
\centering
\caption{Statistics of datasets.}\label{table:data}
{\scalebox{.81}{
\begin{tabular}{l|ccccc}
\hline
~~~~Dataset & NIPS & Kinship & WebKB & AstroPh & Gowalla \\
\hline
Entities & 234 & 104 & 877 & 17,903 & 196,591\\
Positive Links & 1,196 & 415 & 1,608 &391,462 & 1,900,654\\
Sparsity Rate & 2.2\% & 4.1\% & 0.21\% & 0.12\% & 0.0049\%\\
\hline
\end{tabular}}}
\label{table1}
\end{table}

\section{Experiments}\label{sec:Experiments}

We present experimental results to demonstrate the effectiveness of DLFRM on five real datasets as summarized in Table~\ref{table:data}, where {\bf NIPS} contains $234$ authors who have the most coauthor-relationships with others from NIPS $1$-$17$; {\bf Kinship} includes $26$ relationships of $104$ people in the Alyawarra tribe in central Australia; {\bf WebKB} contains $877$ webpages from the CS departments of different universities, where the dictionary has $1,703$ unique words; {\bf AstroPh} contains collaborations between $17,903$ authors of papers submitted to Arxiv Astro Physics in the period from Jan. 1993 to Apr. 2003~\cite{leskovec2007graph}; and {\bf Gowalla} contains $196,591$ people and their friendships on Gowalla social website~\cite{cho2011friendship}. All these real networks have very sparse links.

We evaluate three variants of our model: (1) {\bf DLFRM}: to overcome the imbalance issue, we set $c^+ = 10c^- = c$ as in \cite{Zhu12}, where $c^+$ is the regularization parameter for positive links and $c^-$ for negative links. We use a full asymmetric weight matrix $U$; (2) {\bf stoDLFRM}: the DLFRM model that uses SGLD to sample weight matrix $U$, where the stepsizes are set by $\epsilon_t=a(b+t)^{-\gamma}$ for log-loss and AdaGrad~\cite{duchi2011adaptive} for hinge loss; (3) {\bf diagDLFRM}: the DLFRM that uses a diagonal weight matrix $U$. Each variant can be implemented with the logistic log-loss or hinge loss, denoted by the superscript $l$ or $h$.

We randomly select a development set from training set with almost the same number of links as testing set and choose the proper hyper-parameters, which are insensitive in a wide range. All the results are averaged over $5$ runs with random initializations and the same group of parameters.

\subsection{Results on Small Networks}
We first report the prediction performance (AUC scores) on three relatively small networks. For fair comparison, we follow the previous settings to randomly choose $80\%$ of the links for training and use the remaining $20\%$ for testing. AUC score is the area under the Receiver Operating Characteristic (ROC) curve; higher is better.

\subsubsection{NIPS Coauthorship Prediction}
Table \ref{table:nipsauc} shows the AUC scores on NIPS dataset, where the results of baselines (i.e., LFRM, IRM, MMSB, MedLFRM and BayesMedLFRM) are cited from \cite{Miller09,Zhu12}. We can see that both DLFRM$^l$ and DLFRM$^h$ outperform all other models, which suggests that our exact Gibbs sampling with data augmentation can lead to more accurate models than MedLFRM / BayesMedLFRM that uses the variational approximation methods with truncated mean-field assumptions. The stoDLFRMs obtain comparable results to DLFRMs, which suggests that approximate sampler for $\eta$ using SGLD is  very effective. With SGLD, we can improve efficiency without  sacrificing performance which we will discuss later with Table~\ref{table:aptime}. Furthermore, diagDLFRM$^l$ and diagDLFRM$^h$ also perform well, as they beat all other methods except (sto)DLFRMs. By using a lower dimensional $\eta$ derived from the diagonal weight matrix $U$, diagDLFRM has the advantage of being computationally efficient, as shown in Fig.~\ref{fig:experiments}(d). The good performance of stoDLFRMs and diagDLFRMs suggests that we can use SGLD with a full weight matrix or simply use a diagonal weight matrix on large-scale networks.

\begin{table}[t]
\centering
\caption{AUC on the NIPS coauthorship and Kinship dataset.}
{\scalebox{.84}{\begin{tabular}{l|c|c}
\hline
~~~~~~~~~~Models & NIPS & Kinship \\
\hline
MMSB                &0.8705 $\pm$ $-~~~~~~~~$       &0.9005 $\pm$ 0.0022\\
IRM                 &0.8906 $\pm$ $-~~~~~~~~$  &0.9310 $\pm$ 0.0023\\
LFRM rand           &0.9466 $\pm$ $-~~~~~~~~$  &0.9443 $\pm$ 0.0018\\
LFRM w / IRM        &0.9509 $\pm$ $-~~~~~~~~$  &0.9346 $\pm$ 0.0013\\
MedLFRM             &0.9642 $\pm$ 0.0026       &0.9552 $\pm$ 0.0065\\
BayesMedLFRM        &0.9636 $\pm$ 0.0036       &0.9547 $\pm$ 0.0028\\
\hline
DLFRM$^l$           &$\textbf{0.9812}~\pm$ 0.0013   &$\textbf{0.9650}~\pm$ 0.0032\\
stoDLFRM$^l$           &$\textbf{0.9804}~\pm$ 0.0007   &$\textbf{0.9673}~\pm$ 0.0044 \\
diagDLFRM$^l$       &0.9717 $\pm$ 0.0031            &0.9426 $\pm$ 0.0028 \\
DLFRM$^h$          &\textbf{0.9806} $\pm$ 0.0027   &\textbf{0.9640} $\pm$ 0.0023 \\
stoDLFRM$^h$         &\textbf{0.9787} $\pm$ 0.0012   &\textbf{0.9657} $\pm$   0.0031\\
diagDLFRM$^h$     &0.9722 $\pm$ 0.0021            &0.9440 $\pm$ 0.0038\\
\hline
\end{tabular}}}
\label{table:nipsauc}
\end{table}

\subsubsection{Kinship Multi-relation Prediction}
For multi-relational Kinship dataset, we consider the ``single" setting~\cite{Miller09}, where we infer an independent set of latent features for each relation. The overall AUC is obtained by averaging the results of all relations. As shown in Table~\ref{table:nipsauc}, both (sto)DLFRM$^l$ and (sto)DLFRM$^h$ outperform all other methods, which again proves the effectiveness of our methods. Furthermore, the diagonal variants also obtain fairly good results, close to the best baselines. Finally, the better results by the discriminative methods in general demonstrate the effect of RegBayes on using various regularization parameters to deal with the imbalance issue; Fig.~\ref{fig:experiments}(b) provides a detailed sensitivity analysis.

\subsubsection{WebKB Hyperlink Prediction}
We also examine how DLFRMs perform on WebKB network, which has rich text attributes~\cite{craven1998learning}. Our baselines include: 1) {\bf Katz}: a proximity measure between two entities---it directly sums over all collection of paths, exponentially damped by the path length to count short paths more heavily \cite{Liben-Nowell03}; 2) {\bf Linear SVM}: a supervised learning method using linear SVM, where the feature for each link is a vector concatenating the bag-of-words features of two entities; 3) {\bf RBF-SVM}: SVM with the RBF kernel on the same features as the linear SVM. We use SVM-Light~\cite{svmlight} to train these classifiers; and 4) {\bf MedLFRM}: state-of-the-art methods on learning latent features for link prediction~\cite{Zhu12}. Note that we don't compare with the relational topic models~\cite{Chang:RTM09,Chen13}, whose settings are quite different from ours.
Table~\ref{table:webkbauc} shows the AUC scores of various methods. We can see that: 1) both MedLFRM and DLFRMs perform better than SVM classifiers on raw bag-of-words features, showing the promise of learning latent features for link prediction on document networks; 2) DLFRMs are much better than MedLFRM{\footnote{MedLFRM results are only available when truncation level $<$ 20 due to its inefficiency.}}, suggesting the advantages of using data augmentation techniques for accurate inference over variational methods with truncated mean-field assumptions; and 3) both stoDLFRMs and diagDLFRMs achieve competitive results with faster speed.

\begin{table}[t]
\centering
\caption{AUC scores on the WebKB dataset.}\label{table:webkbauc}
{\scalebox{.82}{\begin{tabular}{l|c}
\hline
~~~~~~~~~~Models ~~~&~~~ WebKB ~~~\\
\hline
~~~Katz            ~~~&~~~ 0.5625  $\pm$ $-~~~~~~~~$ ~~~\\
~~~Linear SVM     ~~~&~~~ 0.6889 $\pm$ $-~~~~~~~~$ ~~~\\
~~~RBF SVM         ~~~&~~~ 0.7132 $\pm$ $-~~~~~~~~$  ~~~\\
~~~MedLFRM         ~~~&~~~ 0.7326 $\pm$ 0.0010       ~~~\\
\hline
~~~DLFRM$^l$     ~~~&~~~ \textbf{0.8039}~$\pm$ 0.0057  ~~~\\
~~~stoDLFRM$^l$    ~~~&~~~ \textbf{0.8044}~$\pm$ 0.0058  ~~~\\
~~~diagDLFRM$^l$  ~~~&~~~ 0.7954 $\pm$ 0.0085          ~~~\\
~~~DLFRM$^h$      ~~~&~~~ \textbf{0.8002}~$\pm$ 0.0073 ~~~\\
~~~stoDLFRM$^h$   ~~~&~~~ \textbf{0.7966}~$\pm$ 0.0013 ~~~\\
~~~diagDLFRM$^h$  ~~~&~~~ 0.7900 $\pm$ 0.0056          ~~~\\
\hline
\end{tabular}}}
\end{table}

\subsection{Results on Large Networks}
We now present results on two much larger networks. As the networks are much sparser, we randomly select $90\%$ of the positive links for training and the number of negative training links is $10$ times the number of positive training links. The testing set contains the remaining $10\%$ of the positive links and the same number of negative links, which we uniformly sample from the negative links outside the training set. This test setting is the same as that in \cite{Kim:2013}.

\begin{figure}[t]
\centering
\includegraphics[height=1.3in, width=3.2in]{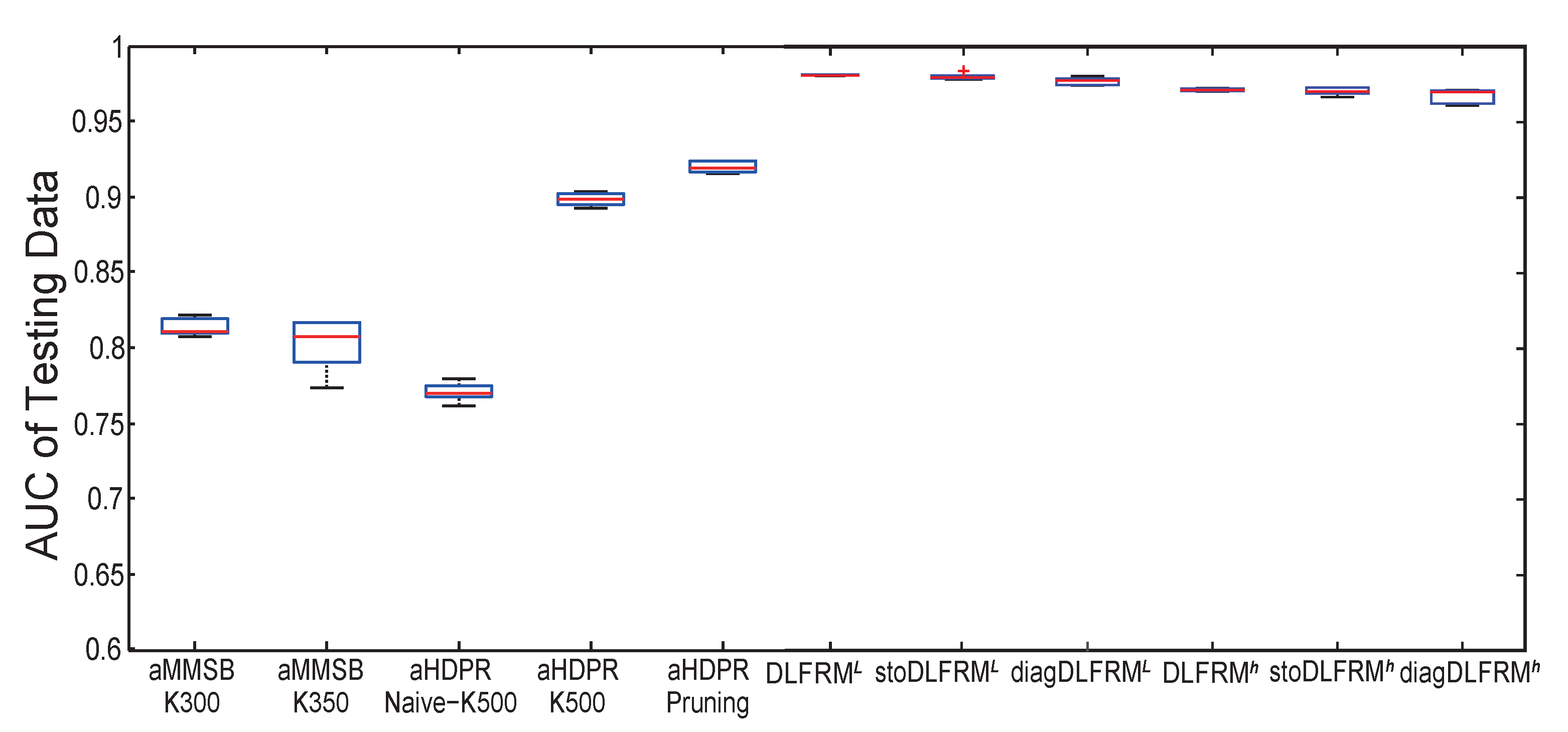}
\caption{AUC scores on the AstroPh dataset.}\label{fig:apauc}
\end{figure}

\begin{table}[t]
\centering
\caption{Split of training time (sec) on AstroPh dataset.}
\scalebox{.8}{\begin{tabu}{X[0.6l]|X[1.2c]|X[1.2c]}
\hline
~~~Models & DLFRM$^l$ &stoDLFRM$^l$  \\
\hline
Sample $Z$ &$16312.0~(25.58\%)$& $32095.9~(95.18\%)$ \\
 Sample $U$ &  $47389.9~(74.32\%)$ &$1516.4~(4.50\%)$\\
 Sample $\lambda$   & $65.7~(0.10\%)$ & $109.0~(0.32\%)$\\
\hline
\end{tabu}}
\label{table:aptime}
\end{table}

\subsubsection{AstroPh Collaboration Prediction}
Fig.~\ref{fig:apauc} presents the test AUC scores, where the results of the state-of-the-art nonparametric models aMMSB (assortative MMSB) and aHDPR (assortative HDP relational model, a nonparametric generalization of aMMSB) are cited from \cite{Kim:2013}. We can see that DLFRMs achieve significantly better AUCs than aMMSB and aHDPR, which again demonstrates that our models can not only automatically infer the latent dimension, but also learn the effective latent features for entities. Furthermore, stoDLRMs and diagDLFRMs show larger benefits on the larger networks due to the efficiency. As shown in Table~\ref{table:aptime}, the time for sampling $U$ is greatly reduced with SGLD. It only accounts for $4.50\%$ of the whole time for stoDLFRM$^l$, while the number is $74.32\%$ for DLFRM$^l$.

\begin{table}[h]
\centering
\caption{AUC scores on Gowalla dataset.}
\scalebox{.8}{\begin{tabu}{X[0.7l]|X[0.9c]|X[1c]}
\hline
~~~~Models & AUC & Time (sec) \\
\hline
CN & 0.8823 $\pm$ $-~~~~~~~~$&12.3 $\pm$ 0.3  \\
Jaccard & 0.8636 $\pm$ $-~~~~~~~~$ &11.7 $\pm$ 0.5 \\
Katz  &0.9145 $\pm$ $-~~~~~~~~$  &8336.9 $\pm$ 306.9\\
\hline
stoDLFRM$^l$ &\textbf{0.9722} $\pm$ 0.0013&220191.4 $\pm$ 4420.2\\
diagDLFRM$^l$ &\textbf{0.9680} $\pm$ 0.0009&7344.5 $\pm$ 943.7\\
\hline
\end{tabu}}
\label{table:gowallaauc}
\end{table}

\subsubsection{Gowalla Friendship Prediction}
\begin{figure}[t]
\centering
\subfigure[]{\includegraphics[height=1.2in, width=1.5in]{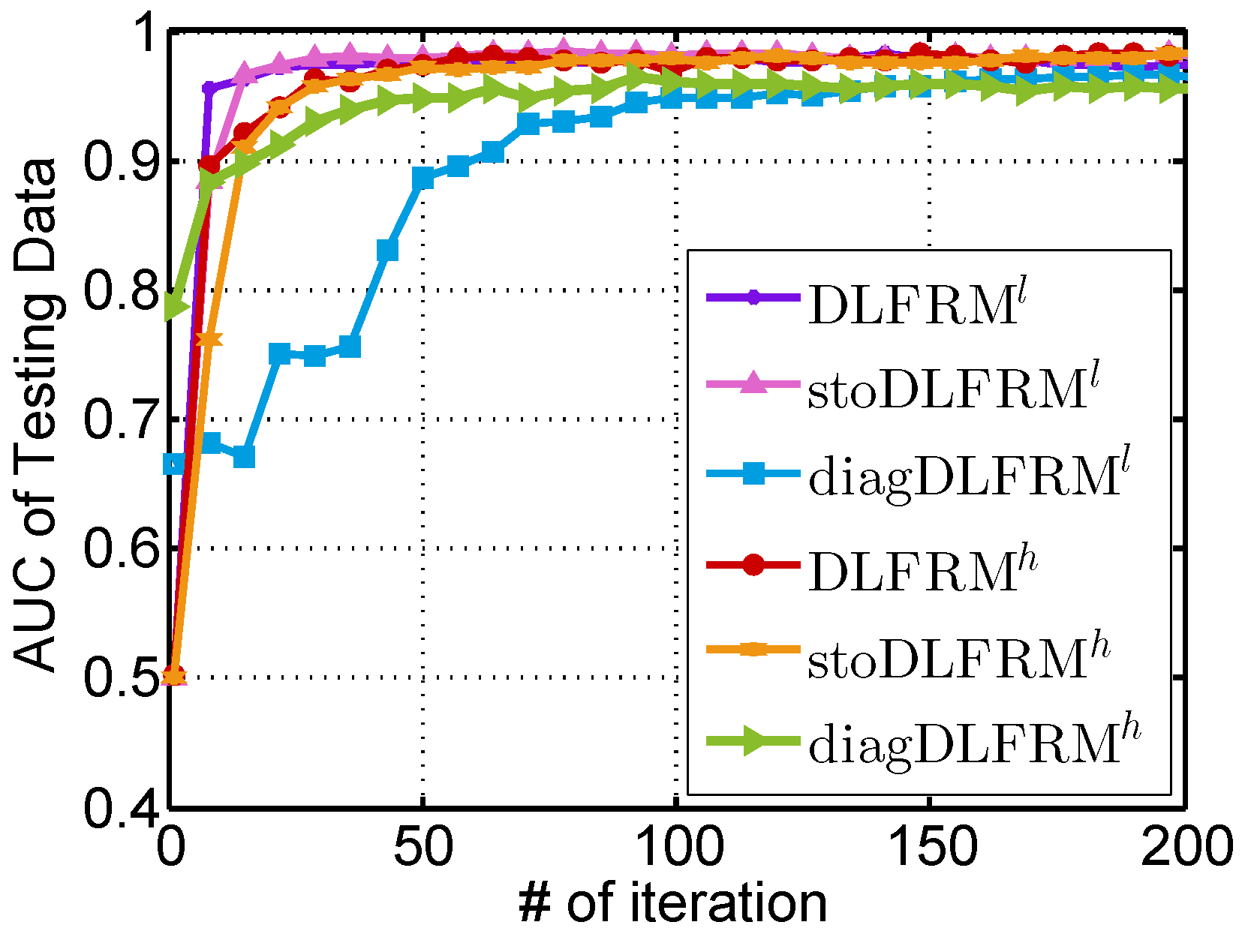}}
\subfigure[]{\includegraphics[height=1.2in, width=1.5in]{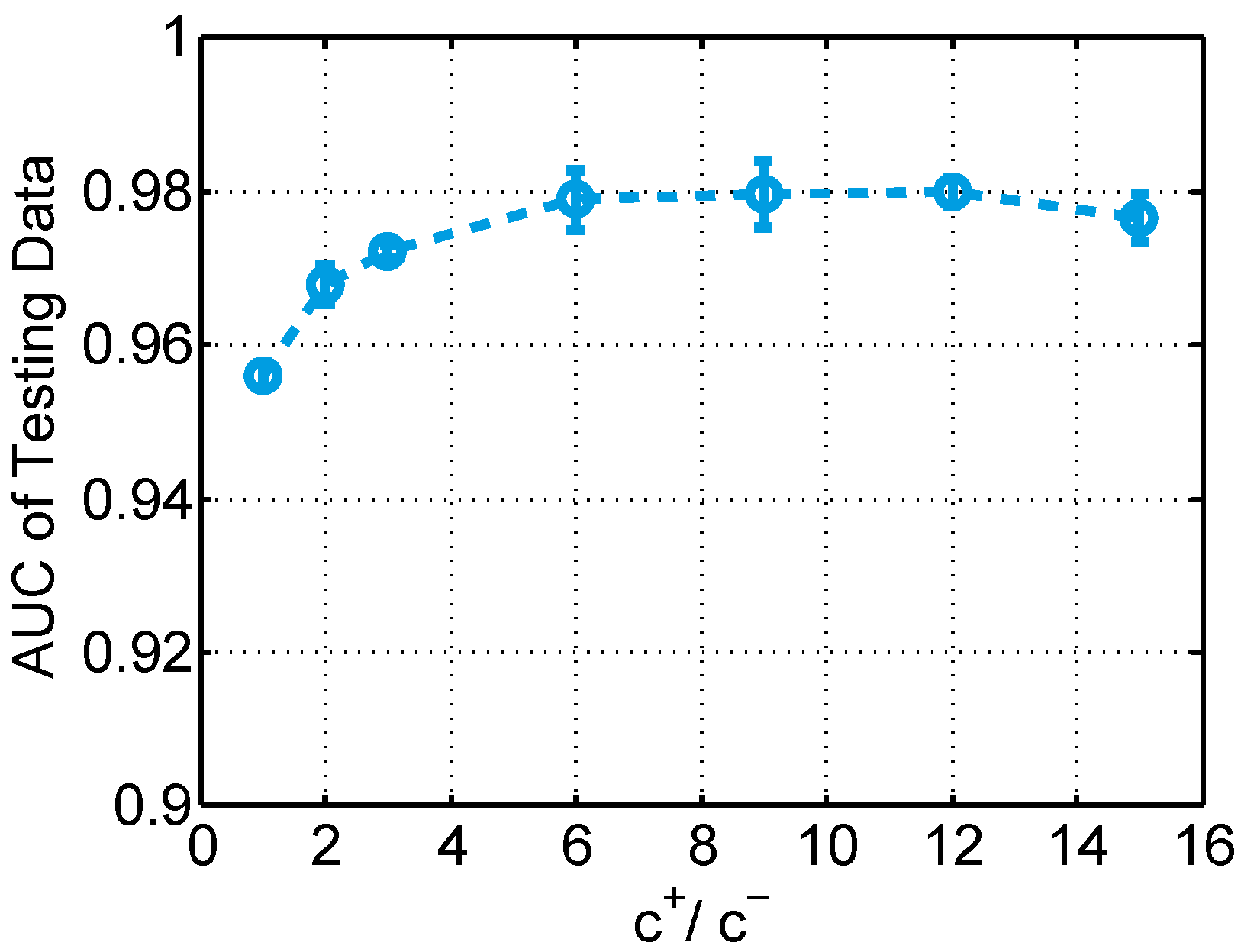}}
\subfigure[]{\includegraphics[height=1.2in, width=1.5in]{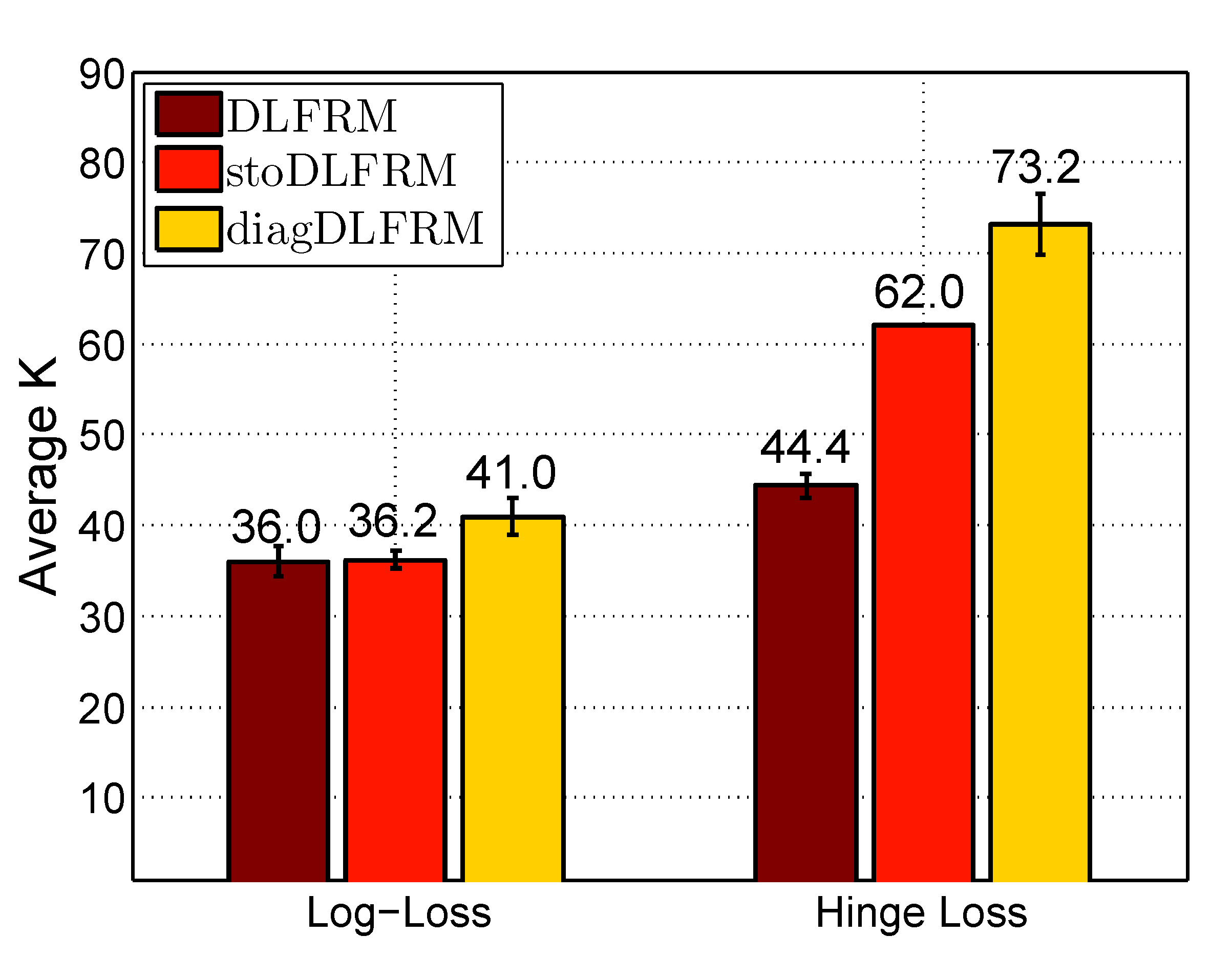}}
\subfigure[]{\includegraphics[height=1.2in, width=1.5in]{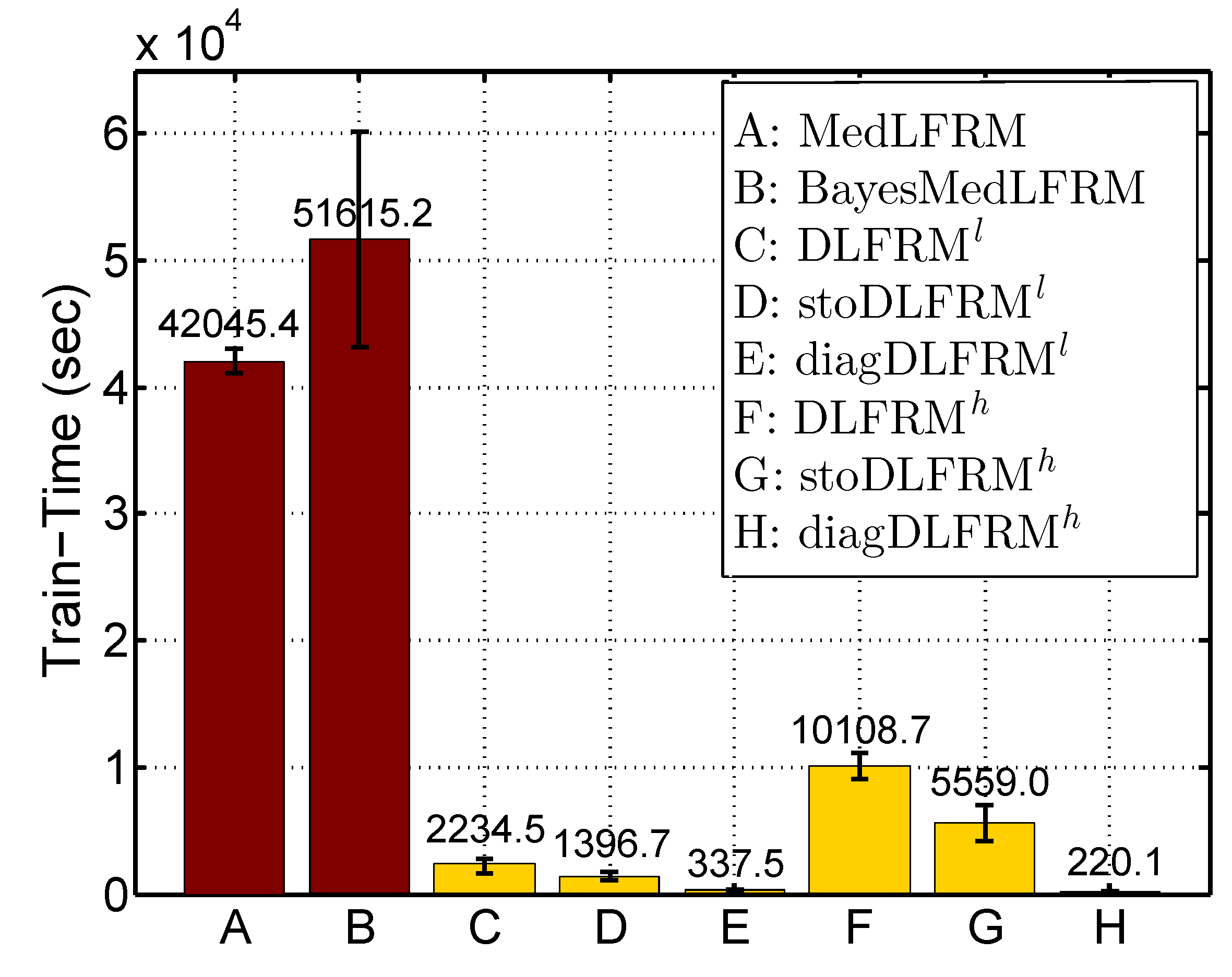}}
\caption{(a) Sensitivity of burn in iterations; (b) Sensitivity of $c^+\! / c^-$ with DLFRM$^l$; (c) Average latent dimension $K$; (d) Training time of various models on NIPS dataset.}
\label{fig:experiments}
\end{figure}

Finally, we test on the largest Gowalla network, which is out of reach for many state-of-art methods, including LFRM, MedLFRM and our DLFRMs without SGLD. Some previous works combine the geographical information of Gowalla social network to analyze user movements or friendships~\cite{cho2011friendship,scellato2011exploiting}, but we are not aware of any fairly comparable results for our setting of link prediction. Here, we present the results of some proximitiy-measure based methods, including common neighbors ({\bf CN}), {\bf Jaccard} coefficient, and {\bf Katz}. As the network is too large to search for all the paths, we only concern the paths that shorter than $4$ for Katz. As shown in previous results and Fig.~\ref{fig:experiments}(d), DLFRMs with logistic log-loss are more efficient and have comparable results of DLFRMs with hinge loss, so we only show the results of stoDLFRM$^l$ and diagDLFRM$^l$. The AUC scores and training time are shown in Table~\ref{table:gowallaauc}. We can see that stoDLFRM$^l$ outperforms all the other methods and diagDLFRM$^l$ obtain competitive results. Our diagDLFRM$^l$ gets much better performance than the best baseline with less time. It shows that our models can also deal with the large-scale networks.

\subsection{Closer Analysis}

We use NIPS network as an example to provide closer analysis. Similar observations can be obtained in larger networks (e.g., AstroPh in Appendix B), but taking longer time to run.

\subsubsection{Sensitivity to Burn-In}
Fig.~\ref{fig:experiments}(a) shows the test AUC scores w.r.t. the number of burn-in steps. We can see that all our variant models converge quickly to stable results. The diagDLFRM$^l$ is a bit slower, but still within $150$ steps. These results demonstrate the stability of our Gibbs sampler.

\subsubsection{Sensitivity to Parameter $c$}
To study how the regularization parameter $c$ handles the imbalance in real networks, {{we change the value of $c^+ \! / c^-$ for DLFRM$^l$ from $1$ to $15$ (with all other parameters selected by the development set)}}; and report AUC scores in Fig.~\ref{fig:experiments}(b). The first point (i.e., $c^+=c^-=1$) corresponds to LFRM with our Gibbs sampler, whose lower AUC demonstrates the effectiveness of a larger $c^+ \! / c^-$ to deal with the imbalance issue. We can see that the AUC score increases when $c^+ \! / c^-$ becomes larger and the prediction performance is stable in a wide range (e.g., $6<c^+ \! / c^-<12$). How large $c^+ \! / c^-$  a network needs depends on its sparsity. A rule of thumb is that the sparser a network is, the larger $c^+ \! / c^-$ it may prefer. The results also show that our setting ($c^+=10c^-$) is reasonable.

\subsubsection{Latent Dimensions}
Fig.~\ref{fig:experiments}(c) shows the number of latent features automatically learnt by variant models. We can see that diagDLFRMs generally need more features than DLFRMs because the simplified weight matrix $U$ doesn't consider pairwise interactions between features. Moreover, DLFRM$^h$ needs more features than DLFRM$^l$, possibly because of the non-smoothness nature of hinge loss. The small variance of each method suggests that the latent dimensions are stable in independent runs with random initializations.

\subsubsection{Running Time}
Fig. \ref{fig:experiments}(d) compares the training time. It demonstrates all our variant models are more efficient than MedLFRM and BayesMedLFRM \cite{Zhu12} that use truncated mean-field approximation. Compared to DLFRM$^l$, DLFRM$^h$ takes more time to get the good AUC. The reason is that DLFRM$^h$ often converges slower (see Fig. \ref{fig:experiments}(a)) with a larger latent dimension $K$ (see Fig. \ref{fig:experiments}(c)). stoDLFRMs are more effective as we have discussed before. diagDLFRMs are much more efficient due to the linear increase of training time per iteration with respect to $K$. The testing time for all the methods are very little, omitted due to space limit.

Overall, DLFRMs improve prediction performance and are more efficient in training, compared with other state-of-the-art nonparametric LFRMs.

\section{Conclusions and Future Work}

We present discriminative nonparametric LFRMs for link prediction, which can automatically resolve the unknown dimensionality of the latent feature space with a simple Gibbs sampler using data augmentation; unify the analysis for both logistic log-loss and hinge loss; and deal with the imbalance issue in real networks. Experimental results on a wide range of real networks demonstrate superior performance and scalability.
For future work, we are interested in developing more efficient algorithms (e.g., using distributed computing) to solve the link prediction problem in web-scale networks.

\section{ Acknowledgments}
 The work was supported by the National Basic Research Program (973 Program) of China (Nos. 2013CB329403, 2012CB316301), National NSF of China (Nos. 61305066, 61322308, 61332007), TNList Big Data Initiative, and Tsinghua Initiative Scientific Research Program (Nos. 20121088071, 20141080934).

\bibliographystyle{aaai}
{\small{
\bibliography{chen-chen}}}

\newpage
\mbox{}
\newpage

\section*{Supplemental Material}
\appendix

\subsection{Appendix  A: The Proof of Lemma 1}
We prove the cases of logistic log-loss and hinge loss in Lemma 1 respectively.
\begin{proof}
For the case with logistic log-loss, we directly follow the data-augmentation strategy from~\cite{Polson2013}. Let $X$ follow a Polya-Gamma distribution, denoted by $X \sim \mathcal{PG}(a, b)$, that is
{\small\setlength\arraycolsep{1pt}\vspace{-.1cm}
\begin{eqnarray}
X=\frac{1}{2\pi^2}\sum_{d=1}^{\infty} \frac{g_d}{(d-1/2)^2+b^2/(4\pi^2)},\vspace{-.1cm}
\end{eqnarray}}where $a>0$ and $b\in \mathcal{R}$ are parameters and each $g_d \sim \mathcal{G}(a, 1)$ is an independent Gamma random variable. The main result of \cite{Polson2013} provides an alternative expression for the form of $\varphi_1$ in Eq.~(\ref{fai1}) by incorporating an augmented variable $\lambda$:
{\small\setlength\arraycolsep{1pt}\vspace{-.1cm}\begin{eqnarray}
\!\!\!\!\!\!\!\varphi_1(\tilde{y}_{ij}|Z_i, Z_j, U)\! =\! \frac{1}{2^c}\!\int_0^{\infty}\!\!\!\!\exp\! \! {\left(\!\kappa_{ij}\omega_{ij}\!-\!\frac{\lambda_{ij}\omega_{ij}^2}{2} \right)}\!\phi(\lambda_{ij}){\rm{d}}  \lambda_{ij},\label{proof1}
\vspace{-.1cm}\end{eqnarray}}where $\kappa_{ij}=c(\tilde{y}_{ij}-\frac{1}{2})$ and $\phi(\lambda_{ij})=\mathcal{PG}(\lambda_{ij};c, 0)$.

For the case with hinge loss, we take the advantage of data augmentation for support vector machines \cite{Polson2011} and $\varphi_2$ in Eq.~(\ref{fai2}) can be represented as a scale mixture of Gaussian distributions:
{\small\setlength\arraycolsep{1pt}\vspace{-.1cm}\begin{eqnarray}
\!\!\!\!\!\varphi_2({y}_{ij}|Z_i, Z_j, U)\!  =\! \!\!\int_0^{\infty} \! \! \!\!\!\frac{1}{\sqrt{2 \pi \lambda_{ij}}} \exp\! {\left( \! -\frac{{(\lambda_{ij}+c\zeta_{ij})}^2}{2\lambda_{ij}} \! \right) } {\rm{d}}  \lambda_{ij},\label{hh1}
\vspace{-.1cm}\end{eqnarray}}where $\zeta_{ij} = \ell - {y}_{ij}\omega_{ij}$ and $\lambda_{ij}$ is the augmented variable.
By reformulating similar terms in Eq.~(\ref{hh1}), we have:
{\small\setlength\arraycolsep{1pt}\vspace{-.1cm}\begin{eqnarray}
\!\!\!\varphi_2({y}_{ij}|Z_i, Z_j, U) && \!\propto\int_0^{\infty}\frac{1}{\sqrt{2 \pi \lambda_{ij}}} \exp\! \Big( \!
-\frac{1}{2}\Big( \frac{c^2\ell^2}{\lambda_{ij}}+\lambda_{ij} \Big)\Big)\nonumber\\
&& \exp\Big({ c{y}_{ij}\Big(1 + \frac{c \ell}{\lambda_{ij}} \Big)\omega_{ij} -\frac{c^2\omega_{ij}^2}{2\lambda_{ij}}
\Big) } {\rm{d}}  \lambda_{ij}\nonumber\\
&&\!\! \propto\!\!\int_0^{\infty} \!\! \exp\Big( \kappa_{ij}\omega_{ij} -\frac{\rho_{ij} \omega_{ij}^2}{2}
\Big)  \phi(\lambda_{ij}){\rm{d}}\lambda_{ij}.
\vspace{-.1cm}\label{proof2}
\end{eqnarray}}where $\kappa_{ij}=c{y}_{ij}(1 + c \ell  \lambda_{ij}^{-1}),~ \rho_{ij} = c^2 \lambda_{ij}^{-1}$ and $\phi(\lambda_{ij})=\mathcal{GIG}(\frac{1}{2}, 1, c^2\ell^2)$. Given the results of Eq.~(\ref{proof1}) and Eq.~(\ref{proof2}), Lemma 1 holds true.
\end{proof}

\subsection{Appendix  B: Closer Analysis on AstroPh dataset}

\begin{figure}[t]\vspace{-3.5mm}
\centering
\subfigure[]{\includegraphics[height=1.1in, width=1.4in]{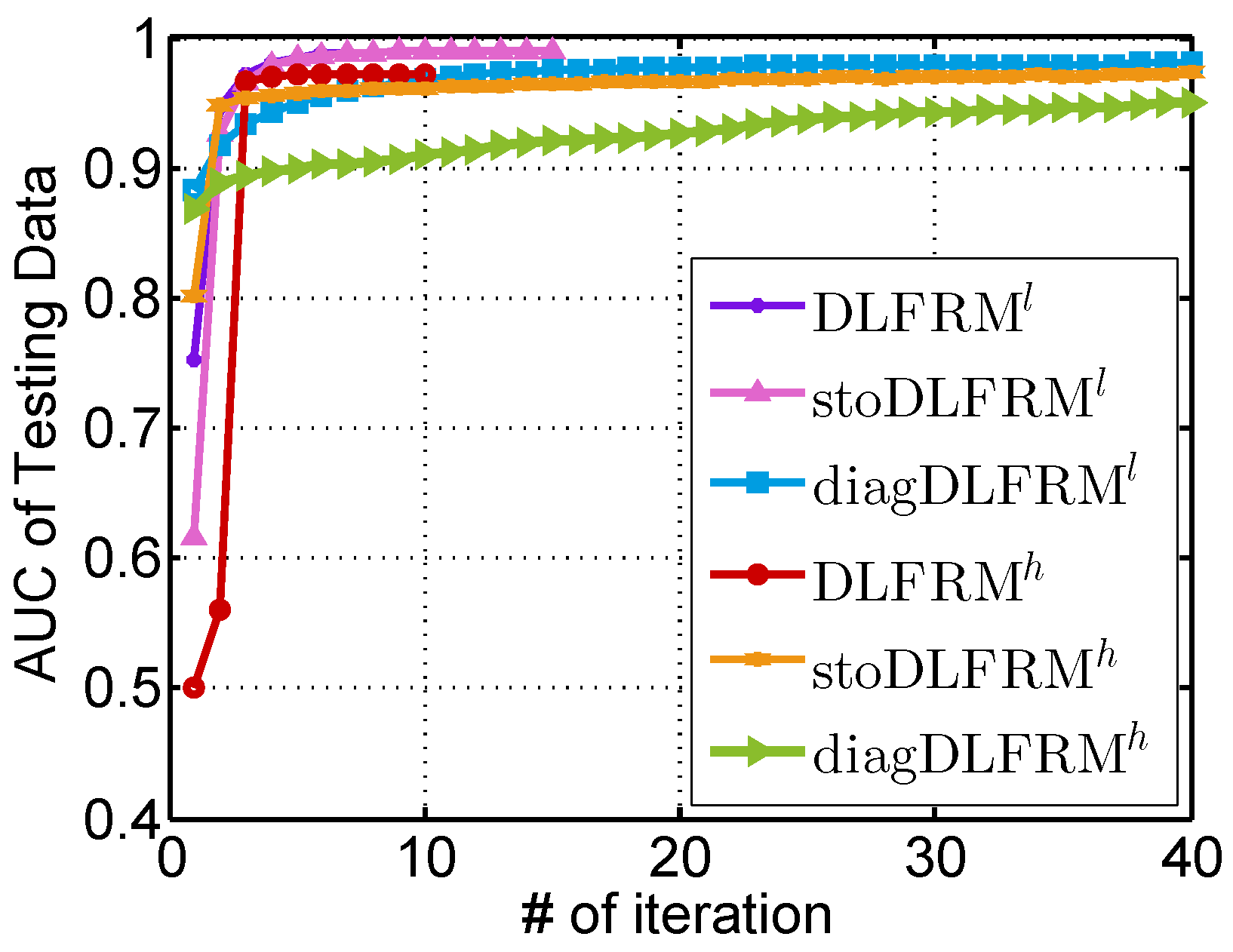}}
\hspace{1mm}
\subfigure[]{\includegraphics[height=1.1in, width=1.4in]{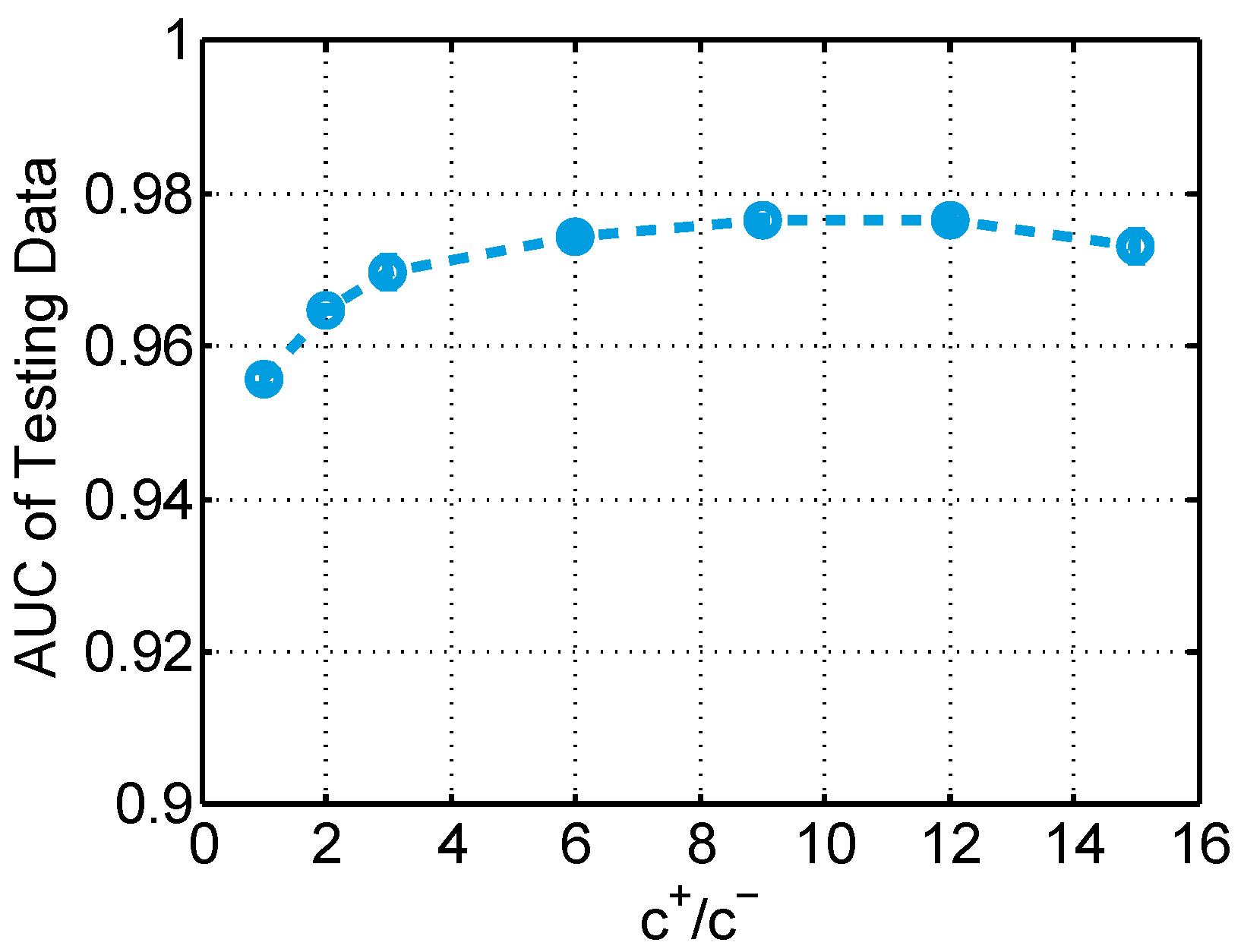}}\\
\vspace{-3mm}
\subfigure[]{\includegraphics[height=1.1in, width=1.4in]{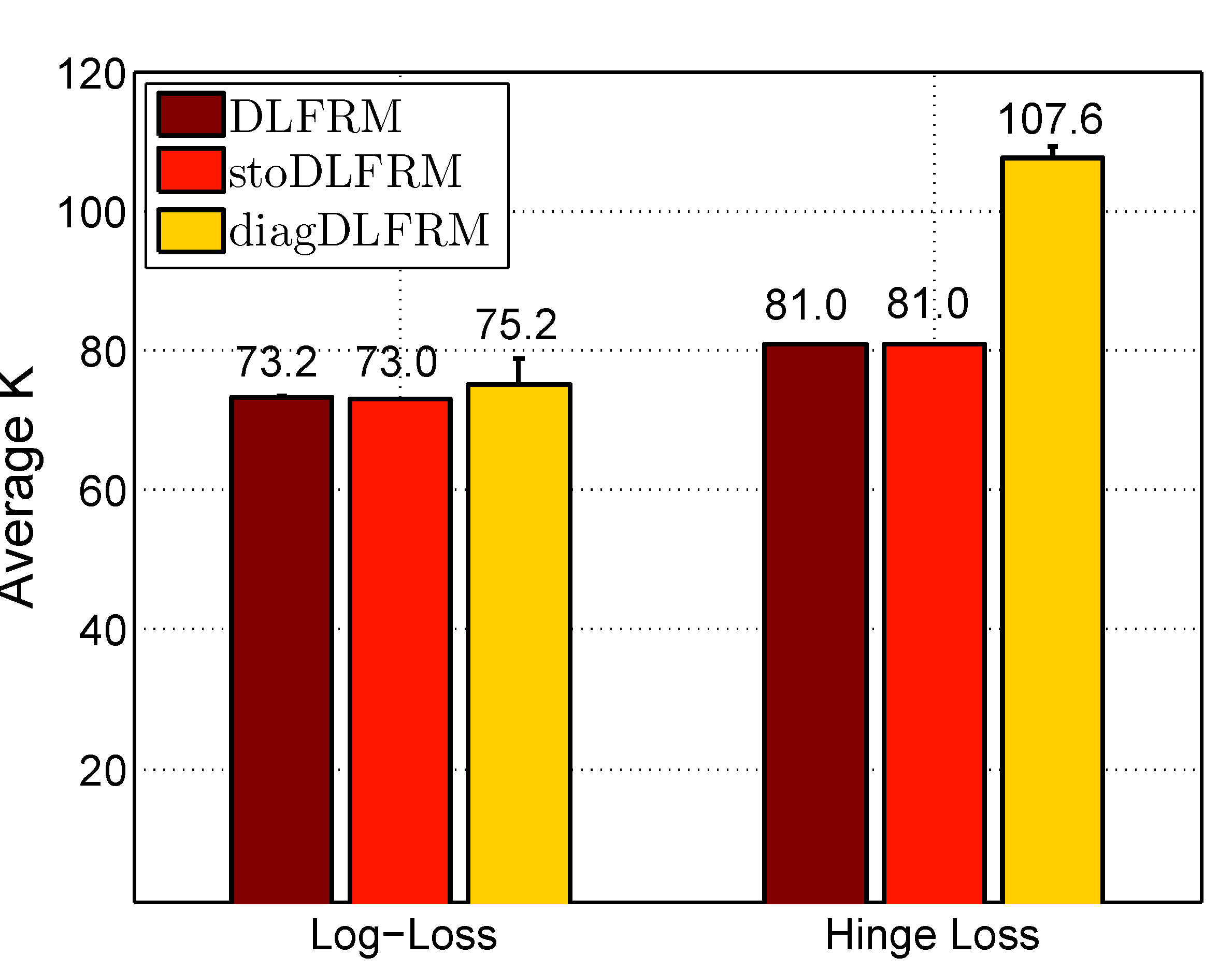}}
\hspace{1mm}
\subfigure[]{\includegraphics[height=1.1in, width=1.4in]{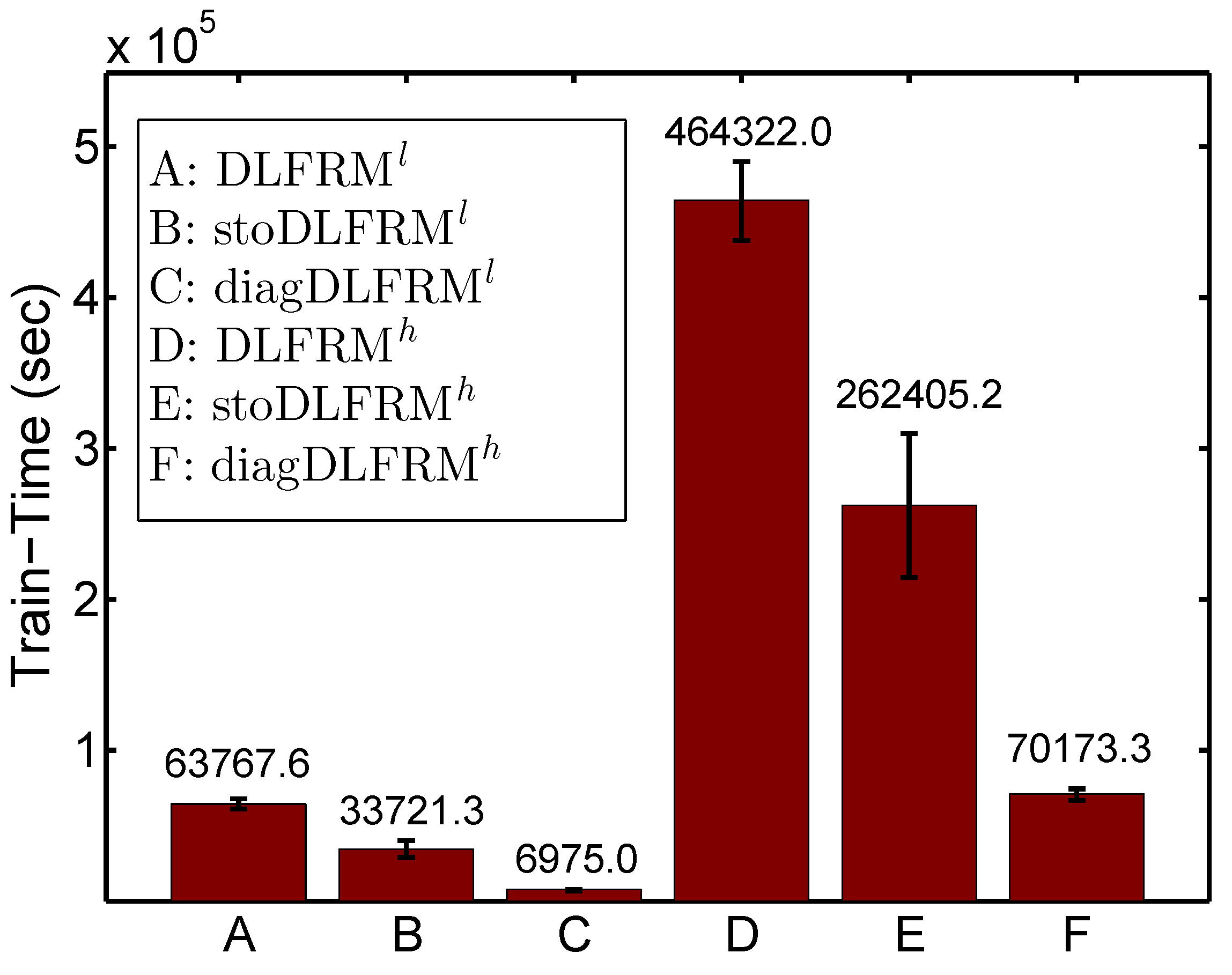}}
\vspace{-4.5mm}
\caption{(a) Sensitivity of burn in iterations; (b) Sensitivity of $c^+\! / c^-$ with diagDLFRM$^l$; (c) Average latent dimension $K$; (d) Training time of various models on AstroPh dataset.}
\label{fig:apexperiments}\vspace{-4.5mm}
\end{figure}

Here, we provide more closer analysis on AstroPh dataset which is much larger than the NIPS dataset.

\subsubsection{Sensitivity to Burn-In}

Fig. \ref{fig:apexperiments}(a) shows the AUC scores on testing data with respect to the number of burn-in steps on AstroPh dataset. We can observe that all our variant models converge quickly to stable results, similar as on NIPS dataset. Our DLFRMs with full weight matrix (e.g., DLFRM$^l$, DLFRM$^h$, stoDLFRM$^l$ and stoDLFRM$^h$) converge quickly within $10$ steps. The diagDLFRMs need more steps to converge, but still within $40$ steps to converge to stable results.
These results demonstrate the stability of our Gibbs sampling algorithm.

\subsubsection{Sensitivity to Parameter $c$}

We analyze how the regularization parameter $c$ handles the imbalance in real networks using diagDLFRM$^l$, which is very efficient (see Fig.~\ref{fig:apexperiments}(d)). Following the settings on NIPS dataset, we change the ratio of $c^+ \! / c^-$ for diagDLFRM$^l$ from $1$ to $15$ with all the parameters selected by the development set. As shown in Fig. \ref{fig:apexperiments}(b), the AUC score increases when $c^+ \! / c^-$ becomes larger and the prediction performance is stable in a wide range (e.g., $6<c^+ \! / c^-<12$). These observations again demonstrate that using a larger $c^+$ than $c^-$  can effectively deal with the imbalance issue and our setting ($c^+=10c^-$) is reasonable.

\subsubsection{Latent Dimensions}

Our variant models take the advantage of nonparametric technique to automatically learn the dimension of the latent features as shown in Fig.~\ref{fig:apexperiments}(c). We can see that diagDLFRMs generally need more features than DLFRMs because the simplified weight matrix $U$ does not consider pairwise interactions between features. Moreover, DLFRM$^h$ needs more features than DLFRM$^l$, possibly because of the non-smoothness nature of hinge loss. The small variance of each method suggests that the latent dimensions are stable in independent runs with random initializations.

\subsubsection{Running Time}

The training time of our variant models on AstroPh dataset is shown in Fig.~\ref{fig:apexperiments}(d).
We can see that for this relatively large network (with tens of thousands of entities and millions of links), the least time we need to obtain the good AUC score is only about $7\times 10^3$ seconds.
As on NIPS dataset, DLFRM$^h$ takes more time for training than DLFRM$^l$ and this phenomenon is more obvious here due to the scalability of the network. The reason is that DLFRM$^h$ often converges slower (see Fig~\ref{fig:apexperiments}(a)) with a larger latent dimension $K$ (see Fig.~\ref{fig:apexperiments}(c)).
As discussed before, stoDLFRMs are more effective.
When a full weight matrix $U$ is used, training time per iteration increases exponentially with respect to $K$. Therefore, diagDLFRMs are much more efficient due to the linear increase of training time per iteration with respect to $K$.

Overall, DLFRMs are stable and improve prediction performance efficiently .

\end{document}